\documentclass[sigconf]{acmart}

\usepackage{algorithm}
\usepackage{algorithmic}
\usepackage{animate} 
\usepackage{multirow}
\usepackage{caption}

\usepackage{multirow}
\usepackage{mathrsfs}
\usepackage{enumitem}
\usepackage{makecell}
\usepackage{tabulary}
\usepackage{pifont}
\usepackage{subfigure}

\usepackage{balance}
\def\BibTeX{{\rm B\kern-.05em{\sc i\kern-.025em b}\kern-.08emT\kern-.1667em\lower.7ex\hbox{E}\kern-.125emX}}

\copyrightyear{2019}
\acmYear{2019}
\acmConference[MM '19]{Proceedings of the 27th ACM International Conference on Multimedia}{October 21--25, 2019}{Nice, France}
\acmBooktitle{Proceedings of the 27th ACM International Conference on Multimedia (MM '19), October 21--25, 2019, Nice, France}
\acmPrice{15.00}
\acmDOI{10.1145/3343031.3350937}
\acmISBN{978-1-4503-6889-6/19/10}

\begin{document}
\fancyhead{}
%
\title{Mocycle-GAN: Unpaired Video-to-Video Translation}
\titlenote{This work was performed at JD AI Research.}
\author{Yang Chen, Yingwei Pan, Ting Yao, Xinmei Tian and Tao Mei}
\affiliation{%
  \institution{{\centering University of Science and Technology of China, Hefei, China}\\{\centering JD AI Research, Beijing, China}}
  }
\email{cheny01@mail.ustc.edu.cn;{panyw.ustc,tingyao.ustc}@gmail.com;xinmei@ustc.edu.cn;tmei@jd.com}

\begin{abstract}
Unsupervised image-to-image translation is the task of translating an image from one domain to another in the absence of any paired training examples and tends to be more applicable to practical applications. Nevertheless, the extension of such synthesis from image-to-image to video-to-video is not trivial especially when capturing spatio-temporal structures in videos. The difficulty originates from the aspect that not only the visual appearance in each frame but also motion between consecutive frames should be realistic and consistent across transformation. This motivates us to explore both appearance structure and temporal continuity in video synthesis. In this paper, we present a new Motion-guided Cycle GAN, dubbed as Mocycle-GAN, that novelly integrates motion estimation into unpaired video translator. Technically, Mocycle-GAN capitalizes on three types of constrains: adversarial constraint discriminating between synthetic and real frame, cycle consistency encouraging an inverse translation on both frame and motion, and motion translation validating the transfer of motion between consecutive frames. Extensive experiments are conducted on video-to-labels and labels-to-video translation, and superior results are reported when comparing to state-of-the-art methods. More remarkably, we qualitatively demonstrate our Mocycle-GAN for both flower-to-flower and ambient condition transfer.
\end{abstract}



\begin{teaserfigure}
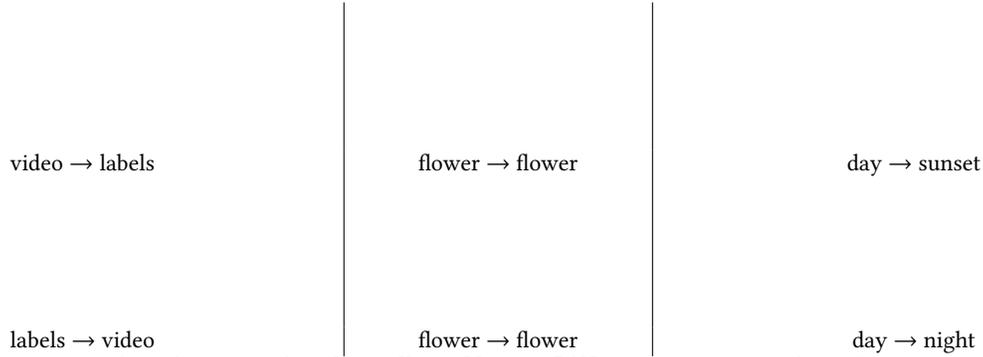

  \begin{tabular}{c| c| c}
        \hspace{0.0cm}\animategraphics[autoplay,loop,width=0.37\textwidth]{12}{1-}{00000}{00019}
        & \animategraphics[autoplay,loop,width=0.21\textwidth]{12}{2-}{00000}{00019}
        &\animategraphics[autoplay,loop,width=0.37\textwidth]{12}{3-}{00000}{00019}
        \\
         \hspace{0.0cm}video $\rightarrow$ labels & flower $\rightarrow$ flower &day $\rightarrow$ sunset\\
         \hspace{0.0cm}\animategraphics[autoplay,loop,width=0.37\textwidth]{12}{4-}{00000}{00019}
        & \animategraphics[autoplay,loop,width=0.21\textwidth]{12}{5-}{00000}{00019}
        &\animategraphics[autoplay,loop,width=0.37\textwidth]{12}{6-}{00000}{00019}
        \\
        \hspace{0.0cm}labels $\rightarrow$ video & flower $\rightarrow$ flower &day $\rightarrow$ night\\
    \end{tabular}
    \vspace{-0.2in}
  \caption{\small Given any two unpaired video collections $X$ and $Y$, our Mocycle-GAN learns to translate videos from source domain $X$ to target domain $Y$. Left: Translation between Game scene videos and segmentation label maps. Center: Translation between time-lapse videos of variant flowers. Right: Translation between Game scene videos under different ambient conditions, e.g., rendering day-light video to the sunset/night environments. The \emph{animated videos} are best viewed via Adobe Acrobat.}
  \label{fig:teaser}
\end{teaserfigure}

%
\maketitle

\section{Introduction}
The development of deep learning has led to a significant surge of research activities for multimedia content generation in multimedia and computer vision community. In between, image-to-image translation is one of the widely studied tasks and the recent advances in Generative Adversarial Networks (GANs) have successfully obtained remarkable improvements on image translation across domains. The achievements of image-to-image translation are on the assumption that a large amount of annotated and matching image pairs are accessible for model training. In practice, nevertheless, the manual labeling of such paired data is cost-expensive and even unrealistic. To address this issue, \cite{liu2016coupled,zhu2017unpaired,kim2017learning,yi2017dualgan} tackle image-to-image translation in an unsupervised manner, which only capitalizes on unpaired data (i.e., two sets of unlabeled images from two domains). In this paper, we go one step further and extend such synthesis from image-to-image to video-to-video, which is referred as an emerging problem of ``unpaired video-to-video translation." It enables a general-purpose video translation across domains in the absence of paired training data, making it flexible to be applied in a variety of video-to-video translation tasks (see Figure \ref{fig:teaser}).

\begin{figure}[!tb]
\vspace{-0.00in}
\centering {\includegraphics[width=0.33\textwidth]{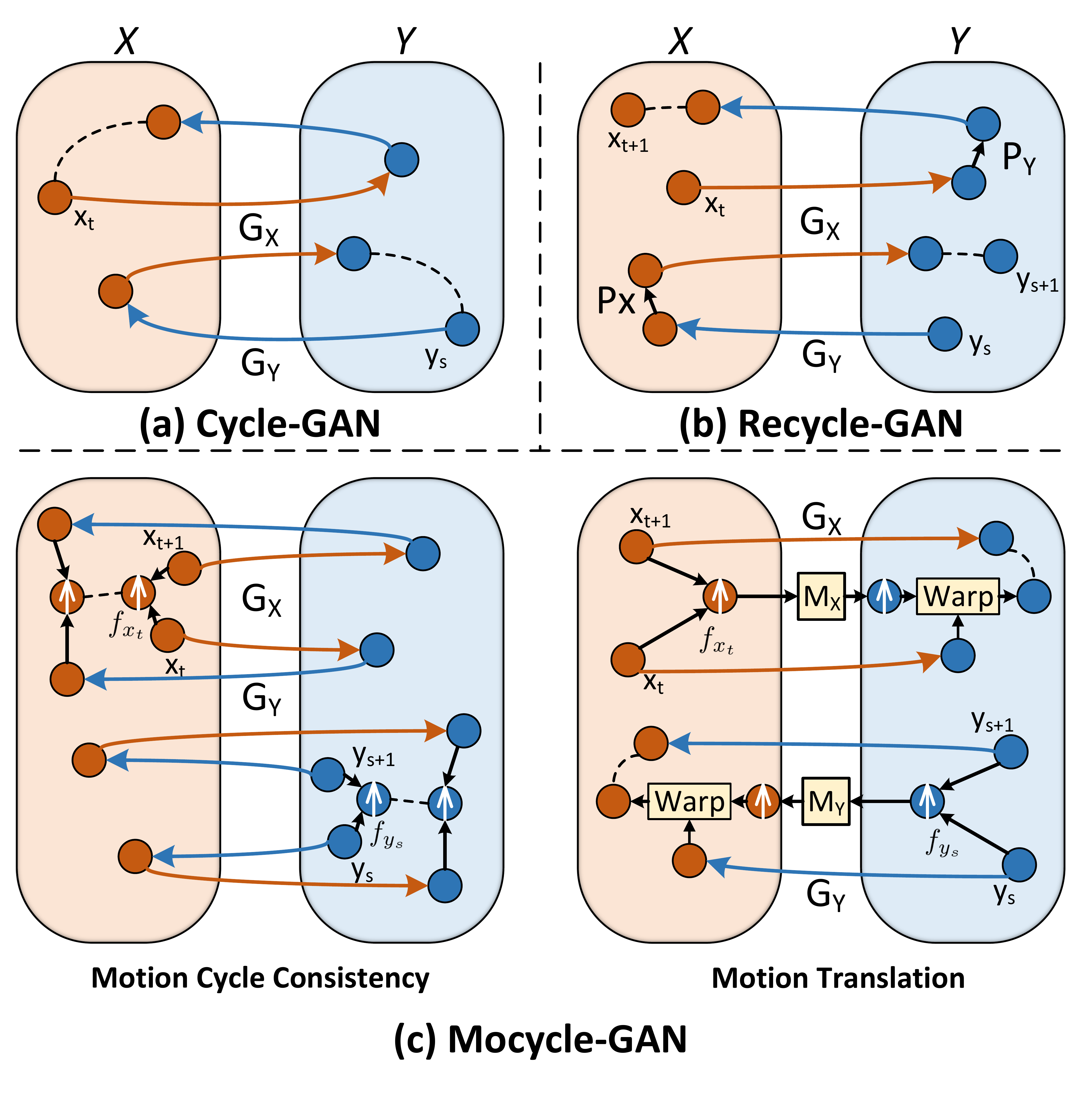}}
\vspace{-0.25in}
\caption{\small Comparison between two unpaired translation approaches and our Mocycle-GAN. (a) \emph{Cycle-GAN} exploits cycle-consistency constraint to model appearance structure for unpaired image-to-image translation. (b) \emph{Recycle-GAN} utilizes temporal predictor ($P_X$ and $P_Y$) to explore cycle consistency across both domains and time for unpaired video-to-video translation. (c) \emph{Mocycle-GAN} explicitly models motion across frames with optical flow ($f_{x_t}$ and $f_{y_s}$), and pursuits cycle consistency on motion that enforces the reconstruction of motion. Motion translation is further exploited to transfer the motion across domains via motion translator ($M_X$ and $M_Y$), strengthening the temporal continuity in video synthesis. Dotted line denotes consistency constraint between its two endpoints.}
\label{fig:figintro2}
\vspace{-0.22in}
\end{figure}

One straightforward way to tackle unpaired video-to-video translation is to capitalize on unpaired image-to-image translation approach, e.g., Cycle-GAN \cite{zhu2017unpaired} (Figure \ref{fig:figintro2}(a)) that enforces an inverse translation for each frame. However, this way only explores visual appearance on frames for video synthesis and will inevitably result in temporal discontinuity when the synthetic frames are deteriorated by flickering artifacts as in video style transfer \cite{chen2017coherent}. This limitation originates from the fact that video is an information-intensive media with complexities along both spatial and temporal dimensions. Such facts motivate and highlight the exploration of both appearance structure and temporal continuity in video synthesis. In this sense, not only the visual appearance in each frame but also motion between consecutive frames are ensured to be realistic and consistent for video translation.

A recent pioneering practice in unpaired video-to-video translation is Recycle-GAN \cite{bansal2018recycle} (Figure \ref{fig:figintro2}(b)). The basic idea is to directly synthesize future frames via temporal predictor to explore cycle consistency across both domains and time. Regardless of the spatio-temporal constraint in Recycle-GAN for enhancing video translation, a common issue not fully studied is the exploitation of motion between consecutive frames, which is well believed to be helpful for video-to-video translation. Instead, we novelly consider the use of motion information for unpaired video-to-video translation from the viewpoint both motion cycle consistency and motion translation, as depicted in Figure \ref{fig:figintro2}(c). The objective of motion cycle consistency constraint is to pursuit cycle consistency on motion between input adjacent frames, which in turn implicitly enforces the temporal continuity between synthetic adjacent frames. In addition, we exploit the constraint of motion translation to further strengthen temporal continuity in synthetic videos via transferring motion across domains. One naive method for enforcing temporal coherence is to warp the synthetic frame with the estimated motion (i.e., optical flow) between input frames to produce the subsequent frame as in \cite{ruder2016artistic,huang2017real}. Nevertheless, this paradigm ignores the occlusions, blur, and appearance variations, e.g., raised by the change of lighting in different domains. As such, the temporal coherence is enforced in a brute-force manner regardless of the scene dynamics in target domain. In comparison, we leverage motion translator to transfer the estimated motion in source domain to target domain, which characterizes the temporal coherence across synthetic frames more tailored to target domain.

\begin{figure*}[!tb]
    \centering {\includegraphics[width=0.88\textwidth]{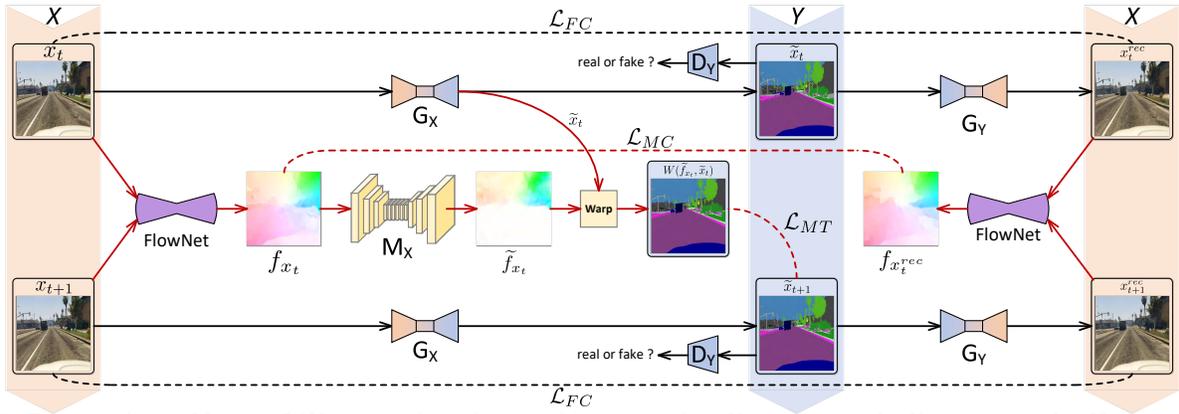}}
    \vspace{-0.20in}
    \caption{\small The overview of Mocycle-GAN for unpaired video-to-video translation ($X$: source domain; $Y$: target domain). Note that here we only depict the forward cycle $X \to Y \to X$ for simplicity. Mocycle-GAN consists of generators ($G_X$ and $G_Y$) to synthesize frames across domains, discriminators ($D_X$ and $D_Y$) to distinguish real frames from synthetic ones, and motion translator ($M_X$) for motion translation across domains. Given two real consecutive frames $x_t$ and $x_{t+1}$, we firstly translate them into the synthetic frames $\widetilde{x}_t$ and $\widetilde{x}_{t+1}$ via $G_X$, which are further transformed into the reconstructed frames $x^{rec}_t$ and $x^{rec}_{t+1}$ through the inverse mapping $G_Y$. In addition, two optical flow $f_{x_t}$ and $f_{x^{rec}_t}$ are obtained by capitalizing on FlowNet to represent the motion before and after the forward cycle. During training, we leverage three kinds of spatial/temporal constrains to explore appearance structure and temporal continuity for video translation: 1) \emph{Adversarial Constraint} ($\mathcal{L}_{Adv}$) ensures each synthetic frame realistic at appearance through adversarial learning; 2) \emph{Frame and Motion Cycle Consistency Constraint} ($\mathcal{L}_{FC}$ and $\mathcal{L}_{MC}$) encourage an inverse translation on both frames and motions; 3) \emph{Motion Translation Constraint} ($\mathcal{L}_{MT}$) validates the transfer of motion across domains in video synthesis. Specifically, the motion translator $M_X$ converts the optical flow $f_{x_t}$ in source to $\widetilde{f}_{x_t}$ in target, which will be utilized to further warp the synthetic frame $\widetilde{x}_t$ to the subsequent frame $W(\widetilde{f}_{x_t}, \widetilde{x}_{t})$. This constraint encourages the synthetic subsequent frame $\widetilde{x}_{t+1}$ to be consistent with the warped version $W(\widetilde{f}_{x_t}, \widetilde{x}_{t})$ in the traceable points, leading to pixel-wise temporal continuity.}
    \label{fig:framework}
    \vspace{-0.19in}
\end{figure*}

By consolidating the idea of exploiting motion information for facilitating unpaired video-to-video translation, we present a novel Motion-guided Cycle GAN (Mocycle-GAN), as shown in Figure \ref{fig:framework}. The whole architecture consists of generators and discriminators under the backbone of standard Conditional GANs, coupled with motion translator for transferring motion across domains. Specifically, the motion information in each domain is estimated in the form of optical flow between consecutive frames. During training, three types of spatial/temporal constrains, i.e., adversarial constraint, cycle consistency on both frame and motion, and motion translation, are devised to explore both the appearance structure and temporal continuity for unpaired video translation. The adversarial constraint discriminates between synthetic and real frames in an adversarial manner, making each synthetic frame realistic at appearance. For the cycle consistency on both frame and motion, it encourages the reconstruction of both appearance structure of frames and temporal continuity in motion. The motion translation constraint transfers the estimated motion from source to target domain via motion translator and then warps the synthetic frame with the transferred motion to the subsequent frame. In this sense, the temporal continuity among synthetic frames in target domain is further strengthened with the guidance from transferred motion. However, unlike in supervised video-video translation, we cannot train the motion translator with paired video data in unpaired scenario. Thus, we optimize the whole architecture in a Expectation Maximization (EM) procedure which iteratively updates generators and discriminators with the three spatial/temporal constrains (E-step), and refines motion translator with an auxiliary motion consistency loss (M-step). Such procedure gradually improves the motion translation as well as the video-to-video translation.

\section{Related Work}\label{sec:RW}
\textbf{Image-to-Image Translation.}
Image-to-image translation aims to learn a mapping function from an input image in one domain to the output image in another domain. The recent advances in GANs \cite{goodfellow2014generative} have inspired the remarkable improvement of this task \cite{isola2017image,zhu2017toward,choi2018stargan,wang2018high}. An early pioneering work \cite{isola2017image} presents a general-purpose solution which leverages Conditional GANs for image-to-image translation. This paradigm enables a variety of graphics tasks, e.g., semantic labels to photo, edges to photo, and photo inpainting. \cite{zhu2017toward} further extends \cite{isola2017image} by encouraging the bijective consistency between the latent and output spaces, leading to more realistic and diverse results. Furthermore, \cite{liu2016coupled,zhu2017unpaired,kim2017learning,yi2017dualgan, yang2018crossing,mejjati2018unsupervised} begin to tackle unsupervised image-to-image translation, i.e., learning to translate images across domains without paired data. In particular, Cycle-GAN \cite{zhu2017unpaired} is devised to learn the mapping function in the absence of paired training data. A cycle consistency loss is utilized to train this mapping coupled with an inverse mapping between the two domains, enforcing the translation to be cycle consistent. Dual GAN \cite{yi2017dualgan} is a concurrent work which also exploits the cycle consistency for unpaired image-to-image translation.

Beyond the still image translation across different domains, our work pursuits its video counterpart by tackling unpaired video-to-video translation in a complex spatio-temporal context. In addition to make each frame realistic, a video translator should be capable of enhancing the temporal coherence among adjacent frames.

\textbf{Video-to-Video Translation.}
Video-to-video translation is a natural extension of image-to-image translation in video domain. Specifically, \cite{wang2018video} is one of the early attempts to tackle video-to-video translation, which integrates a spatio-temporal adversarial objective into conditional GANs. The global and local temporal consistency is exploited in \cite{wei2018video} to ensure the local and global consistency across frames for video-to-video translation. However, the above methods require manual supervision for aligning paired videos across domains, which is extremely expensive and costly to obtain. Inspired from Cycle-GAN \cite{zhu2017unpaired}, \cite{bansal2018recycle} devises Recycle-GAN to facilitate unpaired video-to-video translation. Instead of solely employing spatial constraint for each frame as Cycle-GAN, Recycle-GAN additionally exploits a recurrent temporal predictor to model the dependency between nearby frames, enabling a spatio-temporal constraint (i.e., the recycle consistency) for unpaired video-to-video translation. Video style transfer is another related problem which transfers the style of a reference image to an input video. When directly applying the image style transfer techniques \cite{gatys2016image,ulyanov2016texture,johnson2016perceptual,zhang2018multi,ghiasi2017exploring} to videos, the generated stylized video will inevitably be affected with severe flickering artifacts. As such, to alleviate the flickering artifacts, a number of video style transfer approaches \cite{anderson2016deepmovie,ruder2016artistic,chen2017coherent,huang2017real,gupta2017characterizing,gao2018reconet} are proposed by additionally utilizing temporal constraints to ensure the temporal consistency across frames.

In our work, we also target for an unsupervised solution for video translation. Unlike Recycle-GAN \cite{bansal2018recycle} that directly predicts future frames to enforce the translation to be recycle consistent, our Mocycle-GAN explicitly models the motion across frames with optical flow and pursuits a cycle consistency on motion. Moreover, a motion translator is leveraged to transfer motion in source domain to target domain, aiming to strengthen temporal continuity across synthetic frames with the guidance from transferred motion.

\section{Approach: Mocycle-GAN}

In this paper, we devise Motion-guided Cycle GAN (Mocycle-GAN) architecture to integrate motion estimation into unpaired video translator, exploring both appearance structure and temporal continuity for video translation. The whole architecture of Mocycle-GAN is illustrated in Figure \ref{fig:framework}. We begin this section by elaborating the notation and problem formulation of unpaired video-to-video translation, followed with a brief review of Cycle-GAN with spatial constrain. Then, two kinds of motion-guided temporal constrains, i.e., motion cycle consistency and motion translation, are introduced to further strengthen the temporal continuity. In this sense, both visual appearance in each frame and motion between consecutive frames are ensured to be realistic and consistent across transformation. Finally, the optimization strategy at training along with inference stage are provided.

\subsection{Overview}

\textbf{Notation.}
In unpaired video-to-video translation task, we are given two video collections: $X=\{\bf{x}\}$ in source domain and $Y=\{{\bf{y}}\}$ in target domain, where ${\bf{x}} = \{x_t\}^T_{t=1}$ and ${\bf{y}} = \{y_s\}^S_{s=1}$ denotes the video in source and target domain respectively. $x_t$ and $y_s$ represent the $t$-th frame in source video ${\bf{x}}$ and $s$-th frame in target video ${\bf{y}}$. The goal of this task is to learn two mapping functions between source domain $X$ and target domain $Y$, i.e., $G_X : X \to Y$ and $G_Y : Y \to X$. Here the two mapping functions $G_X$ and $G_Y$ are implemented as generators in Conditional GANs for synthesizing frames. As such, by performing video translation via $G_X$ and $G_Y$, ${\bf{x}}$ and ${\bf{y}}$ are converted as the synthetic videos $\widetilde{\bf{x}} = \{\widetilde{x}_t\}^T_{t=1}$ and $\widetilde{\bf{y}} = \{\widetilde{y}_s\}^S_{s=1}$, where $\widetilde{x}_t=G_X(x_t)$ and $\widetilde{y}_s=G_Y(y_s)$ are synthetic frames. Moreover, one discriminator $D_Y$ is leveraged to distinguish real frames $\{y_s\}$ from synthetic ones $\{\widetilde{x}_t\}$. Similarly, another discriminator $D_X$ distinguishes between $\{x_t\}$ and $\{\widetilde{y}_s\}$. Since we ultimately aim to integrate motion estimation into video translation, we capitalize on off-the-shelf FlowNet \cite{ilg2017flownet} ($\mathcal{F}$) to directly represent the estimated motion between two consecutive frames (e.g., $x_t$ and $x_{t+1}$) as optical flow: $f_{x_t} = \mathcal{F}(x_t, x_{t+1})$. Furthermore, two motion translators, i.e., $M_X$ and $M_Y$, are devised to transfer optical flows across domains. More details about how we conduct motion translation will be elaborated in Section \ref{sec:mo}.

\textbf{Problem Formulation.}
Inspired by the recent success of Cycle-GAN in unpaired image-to-image translation and temporal coherence/dynamics exploration in video understanding \cite{pan2016learning,pan2017video,pan2016jointly,li2018jointly}, we formulate our unpaired video translation model in a cyclic paradigm which enforces the learnt mappings ($G_X$ and $G_Y$) to be cycle consistent on both frame and motion. Specifically, let $x^{rec}_t = G_Y(G_X(x_t))$ and $y^{rec}_s = G_X(G_Y(y_s))$ denotes the reconstructed frame of $x_t$ and $y_s$ in forward cycle and backward cycle, respectively. Hence the frame cycle consistency constraint aims to reconstruct each frame in source and target domain via translation cycle: $x_t \to \widetilde{x}_t \to x^{rec}_t \approx x_t$ and $y_s \to \widetilde{y}_s \to y^{rec}_s \approx y_s$. Besides the preservation of appearance structure in translation cycle via cycle consistency on frame, we additionally pursuit the reconstruction of motion in translation cycle, which enforces the temporal continuity between consecutive frames. As such, the motion cycle consistency constraint is introduced to reconstruct the motion between every two consecutive frames through translation cycle: $f_{x_t} \to  f_{x^{rec}_t} \approx f_{x_t}$ and $f_{y_s} \to f_{y^{rec}_s} \approx f_{y_s}$. In addition, the motion translation constraint is especially devised to exploit motion translation across domains. The transferred motion will be directly utilized to warp the synthetic frame to the subsequent frame, which further strengthens temporal continuity among synthetic frames.

\subsection{Cycle-GAN}
We briefly review Cycle-GAN \cite{zhu2017unpaired} for unpaired translation at frame level. Cycle-GAN is composed of two generators ($G_X$ and $G_Y$) to synthesize frames across domains, and two discriminators ($D_X$ and $D_Y$) for discriminating real frames from synthetic ones, coupled with the adversarial constraint and cycle consistency constraint on frame. The main idea behind Cycle-GAN is to make each frame realistic via adversarial constraint, and encourage the translation cycle-consistent via cycle consistency constraint on frame.

\textbf{Adversarial Constraint.}
As in image/video generation \cite{goodfellow2014generative,pan2017create,qiu2017deep,vondrick2016generating}, the generators and discriminators are adversarially trained in a two-player minimax game mechanism. Specifically, given the real frames ($x_t$ and $y_s$) and the corresponding synthetic frames ($\widetilde{x}_t=G_X(x_t)$ and $\widetilde{y}_s=G_Y(y_s)$), the discriminators are trained to correctly distinguish between real and synthetic frames, i.e., maximizing the adversarial constraint:
\begin{equation}\label{Eq:Eq1}\small
\begin{array}{l}
\mathcal{L}_{Adv} =  \sum\limits_{s}\log D_Y(y_s) + \sum\limits_{t}\log (1 - D_Y(\widetilde{x}_t)) \\
\quad\quad\quad\quad~~~ + \sum\limits_{t}\log D_X(x_t) + \sum\limits_{s}\log (1 - D_X(\widetilde{y}_s)).
\end{array}
\end{equation}
Meanwhile, the generators are learnt to minimize this adversarial constraint, aiming to fool the discriminators with synthetic frames.

\textbf{Frame Cycle Consistency Constraint.}
Moreover, to tackle the unpaired translation, a cycle consistency constraint on each frame is additionally exploited to penalize the difference between the primary input frame $x_t$/$y_s$ and its reconstructed frame $x^{rec}_t = G_Y(G_X(x_t))$/$y^{rec}_s = G_X(G_Y(y_s))$:
\begin{equation}\label{Eq:Eq2}\small
\begin{array}{l}
{\mathcal{L}_{FC}(G_X, G_Y)} = \sum\limits_{t}\left\| x^{rec}_t -  x_t\right\|_1 + \sum\limits_{s}\left\| y^{rec}_s - y_s\right\|_1.
\end{array}
\end{equation}
By minimizing the frame cycle consistency constraint above, the frame translation is enforced to be cycle-consistent, targeting to capture high-level appearance structure across domains.

\subsection{Motion Guided Temporal Constraints}\label{sec:mo}

Unlike Cycle-GAN that only explores appearance structure at frame level, an unpaired video translator should further exploit temporal continuity across frames to ensure both the visual appearance and motion between frames to be realistic and consistent. Existing pioneer in unpaired video translation is Recycle-GAN \cite{bansal2018recycle} that predicts future frames via temporal predictor to enable the cycle consistency across both domains and time, while leaving the inherent motion information unexploited. Here we explicitly model the motion across frames in the form of optical flow throughout the translation. Two temporal constraints, i.e., motion cycle consistency and motion translation, are especially devised to strengthen temporal continuity in synthetic videos with the guidance of motion reconstruction in translation cycle and motion translation across domains.

\textbf{Motion Cycle Consistency Constraint.}
To resolve unpaired scenario of video translation, we go one step further and extend the cycle consistency constraint from single frame in Cycle-GAN to motion between consecutive frames. Formally, given two consecutive frames ($x_{t}$ and $x_{t+1}$) from domain $X$, the forward translation cycle is encouraged to reconstruct the two frames ($x^{rec}_{t}$ and $x^{rec}_{t+1}$) with the consistent optical flow. In other words, the estimated optical flow $f_{x^{rec}_t}$ between $x^{rec}_{t}$ and $x^{rec}_{t+1}$ should be similar to the primary optical flow $f_{x_t}$ between $x_{t}$ and $x_{t+1}$. Similarly, for two consecutive frames ($y_{s}$ and $y_{s+1}$) from domain $Y$, the backward translation cycle is enforced to be cycle-consistent on optical flow: $f_{y_s} \to f_{y^{rec}_s} \approx f_{y_s}$. Accordingly, the motion cycle consistency constraint is defined as the L$_1$ distance between the optical flows before and after the translation cycle:
\begin{equation}\label{Eq:Eq3}\small
\begin{array}{l}
{\mathcal{L}_{MC}(G_X, G_Y)} = \sum\limits_{t}\sum\limits_{i}C^{(i)}_{x_t}\left\|f^{(i)}_{x^{rec}_t} - f^{(i)}_{x_t} \right\|_1 \\
 \quad\quad\quad\quad\quad\quad\quad + \sum\limits_{s}\sum\limits_{i}C^{(i)}_{y_s}\left\| f^{(i)}_{y^{rec}_s}- f^{(i)}_{y_s}\right\|_1,
\end{array}
\end{equation}
where $f^{(i)}_{x_t}$ denotes a 2-dimensional displacement vector for $i$-th pixel in optical flow $f_{x_t}$. As in \cite{huang2017real}, we leverage two visibility masks $C_{x_t}$ and $C_{y_s}$ as weight matrixes, where each pixel $C^{(i)}_{x_t}, C^{(i)}_{y_s} \in \left[ 0,1 \right]$ represents the per-pixel confidence of the pixel $f^{(i)}_{x_t}$ in optical flow $f_{x_t}$: $1$ for traceable pixels by optical flow, and $0$ at occluded regions or near motion boundaries. Accordingly, by minimizing the motion cycle consistency constraint, the video translation is ensured to preserve the motion between real consecutive frames after translation cycle, which in turn implicitly enhances the temporal continuity between synthetic consecutive frames.

\textbf{Motion Translation Constraint.}
The cycle consistency on motion only constraints temporal coherence between synthetic frames in an unsupervised manner, but ignores the straightforward transfer of motion across domains. Nevertheless, the transfer of motion across domains has been seldom exploited for unpaired video translation, possibly because such motion translation needs pairs of optical flows for training, while in the unpaired settings, no paired video data is provided. One naive way to exploit motion across domains for video synthesis is to directly warp the synthetic frame with the source motion into the subsequent frame as in \cite{ruder2016artistic,huang2017real}. This scheme pursuits the motion consistency across domains in a brute-force manner regardless of the scene dynamics in target. Instead, we design a novel motion translator to transfer optical flow from source domain to target domain, which captures temporal coherence tailored to target domain. Such transferred optical flow via motion translator can be further leveraged to guide video synthesis in target domain, pursuing the pixel-wise temporal continuity.

Technically, given the optical flow $f_{x_t}$ between $x_{t}$ and $x_{t+1}$ from domain $X$, the motion translator $M_X$ is utilized to transform the primary optical flow $f_{x_t}$ into the transferred one $\widetilde{f}_{x_t} = M_X(f_{x_t})$ in domain $Y$. Note that motion translators are implemented as paired translator Pix2Pix in \cite{isola2017image}. Each motion translator is constrained with an auxiliary motion consistency loss, aiming to correctly predict the optical flow in the target domain. Here we directly utilize the optical flow $f_{\widetilde{x}_{t}}$ between the corresponding synthetic frames in target domain as the ``pseudo" target optical flow for training motion translator. Similarly, with the input of optical flow $f_{y_s}$ from domain $Y$, another motion translator $M_Y$ produces the transferred optical flow $\widetilde{f}_{y_s} = M_Y(f_{y_s})$ in domain $X$, which is enforced to resemble the ``pseudo" target optical flow $f_{\widetilde{y}_{s}}$ in domain $X$. Thus, the auxiliary motion consistency loss is defined as L$_1$ distance between the transferred optical flow and ``pseudo" target optical flow:
\begin{equation}\label{Eq:Eq4}\small
\begin{array}{l}
{\mathcal{L}_{AM}(M_X, M_Y)} = \sum\limits_{t}\left\| \widetilde{f}_{x_t}-f_{\widetilde{x}_{t}} \right\|_1  + \sum\limits_{s}\left\| \widetilde{f}_{y_s}-f_{\widetilde{y}_{s}} \right\|_1.
\end{array}
\end{equation}
After that, the transferred optical flow $\widetilde{f}_{x_t}$/$\widetilde{f}_{y_s}$ is utilized to further warp the synthetic frame $\widetilde{x}_t$/$\widetilde{y}_s$ to the subsequent frame via bi-linear interpolation, leading to the warped frame $W(\widetilde{f}_{x_t}, \widetilde{x}_{t})$/$W(\widetilde{f}_{y_s}, \widetilde{y}_{s})$ in target domain at time $t+1$. Therefore, we define the motion translation constraint as the L$_1$ distance between the warped frame and the synthetic frame at time $t+1$:
\begin{equation}\label{Eq:Eq5}\small
\begin{array}{l}
{\mathcal{L}_{MT}(G_X, G_Y)} = \sum\limits_{t}\sum\limits_{i}C^{(i)}_{x_t}\left\|W^{(i)}(\widetilde{f}_{x_t}, \widetilde{x}_{t}) - \widetilde{x}^{(i)}_{t+1}\right\|_1 \\
\quad\quad\quad\quad\quad\quad\quad\quad~~ + \sum\limits_{s}\sum\limits_{i}C^{(i)}_{y_s}\left\|W^{(i)}(\widetilde{f}_{y_s}, \widetilde{y}_{s}) - \widetilde{y}^{(i)}_{s+1}\right\|_1.
\end{array}
\end{equation}
This motion translation constraint ensures the synthetic frame to be consistent with the warped version of previous synthetic frame in the traceable points. As such, the pixel-wise temporal continuity among synthetic frames are strengthened.

\subsection{Training and Inference}\label{subsubsec:opt}

\textbf{Optimization.}
The overall training objective of our Mocycle-GAN integrates the adversarial constraint, the cycle consistency constraints on frame and motion, and the motion translation constraint for generators and discriminators, and the auxiliary motion consistency loss for motion translators. During training, we adopt the EM procedure to iteratively optimize motion translators, and generators \& discriminators.
Specifically, in \textbf{E-step}, we fix the parameters in motion translators ($M_X$ and $M_Y$) and update the parameters of generators ($G_X$ and $G_Y$) by minimizing the combined loss of the three spatial/temporal constrains:
\begin{equation}\label{Eq:Eq6}\small
\begin{array}{l}
\mathcal{L}(G_X, G_Y) = \mathcal{L}_{Adv} + \lambda_{FC} \cdot {\mathcal{L}_{FC}}(G_X, G_Y) \\
\quad\quad\quad\quad~~ +  \lambda_{MC} \cdot {\mathcal{L}_{MC}}(G_X, G_Y) + \lambda_{MT} \cdot {\mathcal{L}_{MT}}(G_X,G_Y),
\end{array}
\end{equation}
where $\lambda_{FC}$, $\lambda_{MC}$, and $\lambda_{MT}$ are tradeoff parameters.
Meanwhile, the discriminators ($D_X$ and $D_Y$) are optimized by maximizing the adversarial constraint $\mathcal{L}_{Adv}$ in Eq.(\ref{Eq:Eq1}).
In \textbf{M-step}, we fix the parameters in generators and discriminators, and update motion translators by minimizing the auxiliary motion consistency loss ${\mathcal{L}_{AM}(M_X, M_Y)}$ in Eq.(\ref{Eq:Eq4}). We alternate the E-step and M-step in each training iteration until a convergence criterion is met. The detailed training process of our Mocycle-GAN is given in Algorithm \ref{ag:ag01}. Note that in practice, the generators \& discriminators are pre-trained with the combined loss of adversarial constraint and cycle consistency constraints on frame \& motion. Next, we pre-train the motion translators with the auxiliary motion consistency loss.

\begin{algorithm}[!tb]\scriptsize
\caption{\small The training process of Mocycle-GAN}\label{ag:ag01}
\begin{algorithmic}[1]
    \STATE
        \textbf{Input:} The number of maximum training iteration $N$; Initialize generators ($G_X$, $G_Y$), discriminators ($D_X$, $D_Y$), and motion translators ($M_X$, $M_Y$).
        \FOR{$n=1$ to $N$}
        \STATE
            Fetch input batch with sampled consecutive frame pairs $\{(x_t,x_{t+1}), (y_s, y_{s+1})\}$.
            \FOR {Each consecutive frame pair $(x_t,x_{t+1})$, $(y_s, y_{s+1})$}
            \STATE
            Generate synthetic frames $(\widetilde{x}_t,\widetilde{x}_{t+1}), (\widetilde{y}_s,\widetilde{y}_{s+1})$ and reconstructed frames $ (x^{rec}_t, x^{rec}_{t+1}), (y^{rec}_s, y^{rec}_{s+1})$ via generators ($G_X$, $G_Y$).
            \STATE
            Calculate the corresponding optical flow $f_{x_t}$, $f_{y_s}$, $f_{\widetilde{x}_t}$, $f_{\widetilde{y}_s}$, $f_{x^{rec}_t}$, $f_{y^{rec}_s}$ via FlowNet.
            \STATE
            Produce the transferred flow $\widetilde{f}_{x_t}$ and $\widetilde{f}_{y_s}$ via motion translators ($M_X$, $M_Y$).
            \ENDFOR
            \STATE
                -\textbf{E-step:}
            \STATE
                \hspace{7pt}Fix motion translators ($M_X$, $M_Y$).
            \STATE
                \hspace{7pt}Update generators ($G_X$, $G_Y$) w.r.t loss in Eq.(\ref{Eq:Eq6}).
            \STATE
                \hspace{7pt}Update discriminators ($D_X$, $D_Y$) w.r.t loss in Eq.(\ref{Eq:Eq1}).
            \STATE
                -\textbf{M-step:}
            \STATE
                \hspace{7pt}Fix generators ($G_X$, $G_Y$) and discriminators ($D_X$, $D_Y$).
            \STATE
                \hspace{7pt}Update motion translators ($M_X$, $M_Y$) w.r.t loss in Eq.(\ref{Eq:Eq4}).
        \ENDFOR
\end{algorithmic}
\end{algorithm}

\textbf{Inference.}
After the optimization of our Mocycle-GAN, we can obtain the learnt generator $G_X$ and motion translator $M_X$. During inference, given an input video ${\bf{x}} = \{x_t\}^T_{t=1}$, the simplest way for video translation is to directly employ generator $G_x$ to convert ${\bf{x}}$ into the synthetic video $\widetilde{\bf{x}} = \{\widetilde{x}_t\}^T_{t=1}$ frame-by-frame. An alternative solution is to leverage the warped version of previous synthetic frame based on the transferred optical flow to smooth the output:
\begin{equation}\label{Eq:Eq9}\small
\begin{array}{l}
{\widetilde{x}_{t+1}} = \frac{G_X(x_{t+1}) + W(\widetilde{f}_{x_t}, \widetilde{x}_{t})}{2}.
\end{array}
\end{equation}
However, for fair comparison to other image/video translation approaches, we adopt the simplest single-frame translation without any post processing for evaluation in the experiments.

\section{Experiments}\label{sec:EX}
We empirically verify the merit of our Mocycle-GAN by conducting experiments on four different unpaired video translation scenarios, including video-to-labels, labels-to-video, four ambient condition transfers (day-to-night, night-to-day, day-to-sunset, sunset-to-day) on Viper \cite{richter2017playing} and flower-to-flower on Flower Video Dataset \cite{bansal2018recycle}.

\subsection{Datasets and Experimental Settings}
\textbf{Viper} is a popular visual perception benchmark to facilitate both low-level and high-level vision tasks, e.g., optical flow and semantic segmentation. It consists of videos from a realistic virtual world (i.e. GTA gameplay), which are collected while driving, riding and walking in diverse ambient conditions (day, sunset, snow, rain, and night). Each frame (resolution: $1920 \times 1080$) is annotated with pix-level labels, i.e., segmentation label map. Following \cite{bansal2018recycle}, we split 77 videos under diverse environmental conditions into 57 for training and 20 for testing. For video-to-labels and labels-to-video, we evaluate the translations between videos and segmentation label maps. For ambient condition transfers, we consider the translation across different ambient conditions: day $\leftrightarrow$ night and day $\leftrightarrow$ sunset.

\textbf{Flower Video Dataset} is a recent released dataset for video translation. This dataset includes the time-lapse videos which depict the blooming or fading of various flowers but without any sync. The resolution of each video is 256 $\times$ 256. For flower-to-flower, we evaluate the translation between different types of flowers, aiming to align the high-level semantic content among them, e.g., the two flowers simultaneously bloom or fade at the same pace.

\textbf{Implementation Details.} We mainly implement our Mocycle-GAN on Pytorch \cite{paszke2017automatic} architecture. For generators, we follow the settings of \cite{zhu2017unpaired, bansal2018recycle}, and adopt the encoder-decoder architecture \cite{johnson2016perceptual}. In particular, each generator is composed of two convolution layers (stride: 2) for down-sampling, six residual blocks \cite{he2016deep}, and two deconvolution layers for up-sampling. Each discriminator is built as the $70 \times 70$ PatchGAN in \cite{isola2017image}. For motion translators, we adopt the similar architecture of generator by modifying the input and output channel as 2, which enables the translation of optical flow across domains. In all experiments, we set the tradeoff parameters in Eq.(\ref{Eq:Eq6}) as $\lambda_{FC} = 10$, $\lambda_{MC} = 10$, and $\lambda_{MT} = 10$. During training, the batch size is set as 1. Adam \cite{kingma2014adam} is utilized to optimize the parameters in generators, discriminators and motion translators with the initial learning rate of 0.0002, 0.0002 and 0.0001, respectively.

\begin{table}[!tb]\scriptsize
\centering
\caption{\small Segmentation score (\%) of our Mocycle-GAN and other methods for video-to-labels translation on Viper.}
\setlength{\tabcolsep}{2.8pt}
\def\arraystretch{1.2}
\vspace{-0.15in}
\label{table:vid2seg viper}
\begin{tabular}{@{}c| l| c c c c c c }
\Xhline{2\arrayrulewidth}
Criterion& Approach  & day  &   sunset & rain &  snow & night &  all \\ \hline\hline
\multirow{5}*{\textbf{MP}} & Cycle-GAN \cite{zhu2017unpaired}  &46.0  &68.7	 & 41.1	 &39.2   &32.2    &40.2	\\  	
& Recycle-GAN \cite{bansal2018recycle} 	 		                                &53.0  &	75.6	 & 51.6	 &55.5   &39.7    &58.8	\\  	
& Recycle-GAN$_{cmb}$ \cite{bansal2018recycle}			            &54.7  &	76.3	 & 51.0	 &57.0   &44.7    &60.1	\\
& Cycle-GAN$_{SF}$ \cite{huang2017real}	             &55.2  &77.1     &49.9    &59.6  &42.2     &62.3  \\
& Mocycle-GAN  &\textbf{64.2}  &	\textbf{82.1}	 & \textbf{67.0}	 &\textbf{66.1}   &\textbf{64.5}    &\textbf{64.9}	\\ \hline

\multirow{5}*{\textbf{AC}} & Cycle-GAN \cite{zhu2017unpaired} &12.0  &	13.1	 & 5.1	 &9.5   &4.9    &9.6	\\  		
& Recycle-GAN \cite{bansal2018recycle} 	  		                                &13.5  &	16.8	 & 9.9	 &11.0   &8.4    &14.4	\\  	
& Recycle-GAN$_{cmb}$ \cite{bansal2018recycle}				        &15.3  &	15.7	 & 10.9	 &11.3   &10.2    &14.9	\\  	
& Cycle-GAN$_{SF}$ \cite{huang2017real}	                &16.2   &17.0  &10.7 &13.0  &9.7  &16.1\\
& Mocycle-GAN   &\textbf{20.5}  &	\textbf{23.0}	 & \textbf{18.4}	 &\textbf{17.8}   &\textbf{16.4}    &\textbf{17.7}	\\ \hline

\multirow{5}*{\textbf{IoU}} & Cycle-GAN \cite{zhu2017unpaired}  &7.4  &	9.9	 & 3.1	 &5.8   &2.9    &5.3	\\  		
& Recycle-GAN \cite{bansal2018recycle} 	 		                                &9.4  &	13.1	 & 6.6	 &7.8   &5.2    &10.5	\\  	
& Recycle-GAN$_{cmb}$ \cite{bansal2018recycle}			            &10.8  &	12.4	 & 6.8	 &8.1   &6.4    &11.0	\\  	
& Cycle-GAN$_{SF}$ \cite{huang2017real}	                &11.6   &13.4      &6.3     &9.0   &6.4    &11.0\\
& Mocycle-GAN   &\textbf{15.2}  &	\textbf{18.1}	 & \textbf{11.9}	 &\textbf{12.3}   &\textbf{11.6}    &\textbf{13.2}	\\ \Xhline{2\arrayrulewidth}
\end{tabular}
\vspace{-0.2in}
\end{table}

\textbf{Evaluation Metrics.}
For video-to-labels translation, as in \cite{zhu2017unpaired, bansal2018recycle}, we adopt three standard segmentation metrics in \cite{long2015fully} for evaluation, i.e., Mean Pixel Accuracy (\textbf{MP}), Average Class Accuracy (\textbf{AC}), and Intersection-Over-Union (\textbf{IoU}). For labels-to-video translation, we follow \cite{zhu2017unpaired, bansal2018recycle} and report the \textbf{FCN score} on target domain. FCN score represents the quality of synthetic frames according to an off-the-shelf semantic segmentation network. Specifically, we pre-train a fully-convolutional network, i.e., FCN \cite{long2015fully}, on Viper. Next, the FCN model is utilized to predict the segmentation label map for each synthetic frame. By comparing the predicted segmentation label map against the ground-truth labels, we can obtain the FCN scores with regard to the three standard segmentation metrics described above (i.e., MP, AC, and IoU). The intuition is that the higher the FCN scores, the more realistic the synthetic frames at appearance.

\textbf{Compared Approaches.}
We include the following state-of-the-art unpaired translation methods for performance comparison: $(1)$ \textbf{Cycle-GAN}\cite{zhu2017unpaired} is an unpaired image translator that pursuits an inverse translation only at frame level. $(2)$ \textbf{Recycle-GAN}\cite{bansal2018recycle} leverages a recurrent temporal predictor to generate future frames and pursues a new cycle consistency (i.e. recycle loss) across domains and time for unpaired video-to-video translation. $(3)$ \textbf{Recycle-GAN}$_{cmb}$ \cite{bansal2018recycle} is an upgraded version of Recycle-GAN by combining recycle loss in Recycle-GAN and cycle loss in Cycle-GAN for training video translator. $(4)$ \textbf{Cycle-GAN}$_{SF}$ remoulds a state-of-the-art video style transfer approach \cite{huang2017real} for unpaired video translation by equipping its short temporal constraint with the cycle loss in Cycle-GAN. The basic idea of the short temporal constraint is to directly warp the synthetic frame with the source motion into the subsequent frame, aiming to enforce the pixel-wise temporal consistency. $(4)$ \textbf{Mocycle-GAN} is the proposal in this paper. Please note that for fair comparison, all the baselines and our Mocycle-GAN utilize the same architecture for generators and discriminators.

\begin{table}[!tb]\scriptsize
\centering
\caption{\small FCN score (\%) of our Mocycle-GAN and other methods for labels-to-video translation on Viper.}
\setlength{\tabcolsep}{2.8pt}
\def\arraystretch{1.2}
\vspace{-0.15in}
\label{table:seg2vid viper}
\begin{tabular}{@{}c |l| c c c c c c }
\Xhline{2\arrayrulewidth}
Criterion& Approach  & day  &   sunset & rain &  snow & night &  all \\
\hline\hline
\multirow{6}*{\textbf{MP}} & Cycle-GAN \cite{zhu2017unpaired}		                    &36.3  &	48.7	 & 23.7	 &40.0   &22.8    &37.9	\\  	
& Recycle-GAN \cite{bansal2018recycle} 	 		                                &37.5  &	53.9	 & 27.4	 &42.7   &23.6    &41.3	\\  	
& Recycle-GAN$_{cmb}$ \cite{bansal2018recycle}				            &37.0  &	54.4	 & 27.6	 &40.8   &26.6    &43.5	\\  	
& Cycle-GAN$_{SF}$ \cite{huang2017real}	                &38.7 &57.0 &25.2 &42.1 &24.4  &44.6 \\
& Mocycle-GAN                      &\textbf{42.1}  &\textbf{61.2}	 & \textbf{34.6}	 &\textbf{48.1}   &\textbf{30.5}    &\textbf{47.6}	\\
\hline

\multirow{6}*{\textbf{AC}} & Cycle-GAN \cite{zhu2017unpaired}		                    &10.7  &	15.3	 & 9.1	 &11.4   &10.0    &10.2	\\  		
& Recycle-GAN \cite{bansal2018recycle} 	  		                                &12.3  &	14.9	 & 10.0	 &11.5   &11.1    &12.0	\\  	
& Recycle-GAN$_{cmb}$ \cite{bansal2018recycle}				        &12.7  &	15.6	 & 10.1	 &12.0   &11.8    &12.2	\\  	
& Cycle-GAN$_{SF}$ \cite{huang2017real}	            &13.2  &15.4  &9.7  &13.0 &10.4 &12.9 \\
& Mocycle-GAN                                                      &\textbf{15.4}  &	\textbf{17.6}	 & \textbf{12.6}	 &\textbf{14.9}   &\textbf{16.5}    &\textbf{16.0}	\\
\hline

\multirow{6}*{\textbf{IoU}} & Cycle-GAN\cite{zhu2017unpaired}		                    &7.4  &	9.2	 & 4.7	 &6.2   &4.5    &6.1	\\  		
& Recycle-GAN \cite{bansal2018recycle} 	 		                                &8.1  &	10.0	 & 5.5	 &6.9   &4.7    &6.7	\\  	
& Recycle-GAN$_{cmb}$ \cite{bansal2018recycle}			            &8.3  &	10.2	 & 5.5	 &6.9   &5.6    &7.0	\\  	
& Cycle-GAN$_{SF}$ \cite{huang2017real}	            &8.0 &10.4 &5.0 &7.0 &5.3 &7.4 \\
& Mocycle-GAN                   &\textbf{9.7}  &	\textbf{11.9}	 & \textbf{7.5}	 &\textbf{8.8}   &\textbf{7.7}    &\textbf{10.1}	\\
\Xhline{2\arrayrulewidth}
\end{tabular}
\vspace{-0.23in}
\end{table}

\subsection{Performance Comparison and Analysis}

\begin{figure*}[!tb]
    \centering {\includegraphics[width=0.9\textwidth]{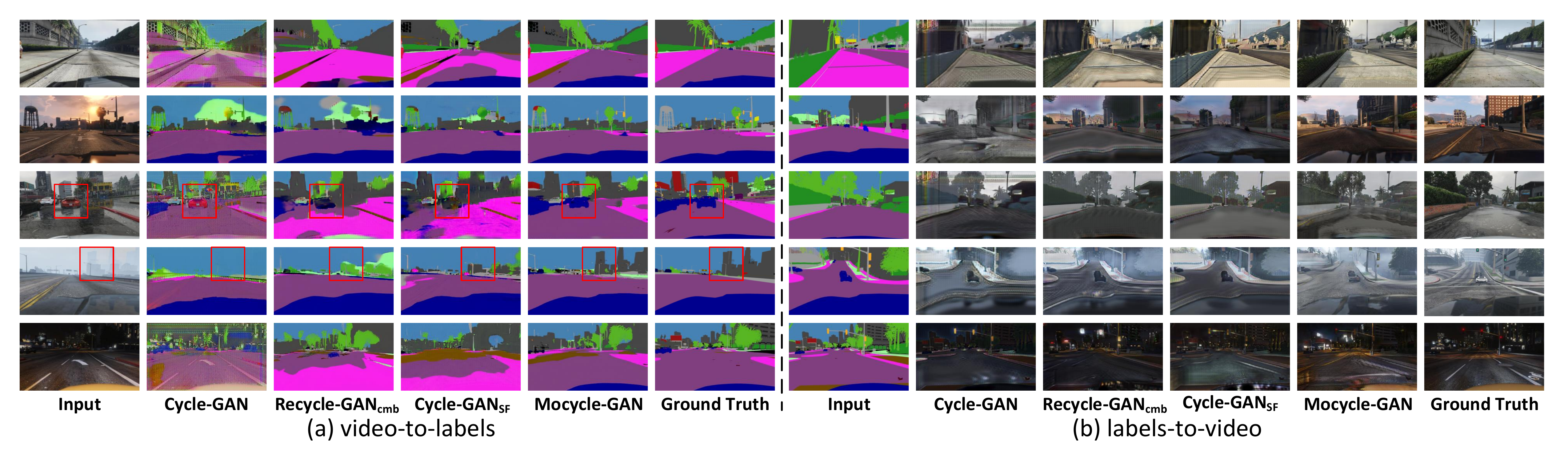}}
    \vspace{-0.30in}
    \caption{\small Examples of (a) video-to-labels and (b) labels-to-video results in Viper dataset under various ambient conditions. The original inputs, the output results by different models, and the ground truth outputs are given.}
    \vspace{-0.18in}
    \label{fig:viper}
\end{figure*}

\textbf{Evaluation on Video-to-Labels.} In this scenario, the video translator takes a game scene video as input and outputs the corresponding segmentation label maps. The performance comparisons of different models for video-to-labels translation task are summarized in Table \ref{table:vid2seg viper}. Overall, the results across three segmentation metrics consistently indicate that our proposed Mocycle-GAN obtains better performances against state-of-the-art techniques. The results generally highlight the key advantage of exploring motion information for unpaired video translation, enforcing the synthetic videos to be both realistic at appearance and temporal continuous across frames. Specifically, by encouraging the cycle consistency across domains and time via a spatio-temporal constraint, Recycle-GAN exhibits better performance than Cycle-GAN that only pursuits cycle consistency at frame level. Moreover, by simultaneously utilizing the spatial constraint in Cycle-GAN and spatio-temporal constraint in Recycle-GAN, Recycle-GAN$_{cmb}$ further boosts up the performances. Different from Recycle-GAN$_{cmb}$ that enforces temporal coherence via future frame prediction, Cycle-GAN$_{SF}$ encourages pixel-wise temporal consistency by directly warping the synthetic frame with source optical flow, and achieves better performances. This confirms the effectiveness of modeling motion information in video synthesis. Nevertheless, the performances of Cycle-GAN$_{SF}$ are still lower than our Mocycle-GAN which further strengthens temporal continuity via motion cycle consistency and motion translation across domains.

\begin{table}[]\scriptsize
\centering
\caption{\small Ablation study for each design (i.e., Motion Cycle Consistency (MC) and Motion Translation (MT)) in Mocycle-GAN for video-to-labels on Viper.}
\setlength{\tabcolsep}{3.98pt}
\def\arraystretch{1.2}
\vspace{-0.15in}
\label{table:ablation-vid2seg}
\begin{tabular}{@{}c| l| c c |c c c c c c }
\Xhline{2\arrayrulewidth}
Criterion& Approach &MC  & MT &day  &   sunset & rain &  snow & night &  all \\ \hline\hline

\multirow{3}*{\textbf{MP}} & Cycle-GAN + MC		&$\surd$ &~    &60.2  &	81.1	 & 61.3	 &63.0   &50.7    &63.1	\\
& Cycle-GAN + MT				  &~ &$\surd$        &62.5  &	81.5	 & 65.4	 &64.6   &63.0    &63.0	\\
& Mocycle-GAN                      &$\surd$ &$\surd$                                &\textbf{64.2}  &\textbf{82.1}		 & \textbf{67.0}	 &\textbf{66.1}   &\textbf{64.5}    &\textbf{64.9}	\\ \hline

\multirow{3}*{\textbf{AC}} & Cycle-GAN + MC	   &$\surd$ &~       &18.2  &	21.3	 & 15.7	 &14.4   &12.2    &17.4	\\
 & Cycle-GAN + MT   			&~ &$\surd$            &19.3  &	21.4	 & 17.6	 &17.4   &16.1    &17.2	\\	
& Mocycle-GAN                                &$\surd$ &$\surd$                      &\textbf{20.5}  &	\textbf{23.0}	 & \textbf{18.4}	 &\textbf{17.8}   &\textbf{16.4}    &\textbf{17.7}	\\ \hline

\multirow{3}*{\textbf{IoU}} & Cycle-GAN + MC	   &$\surd$ &~    &13.3  &	16.9	 & 9.6	 &10.4   &7.9    &12.9	\\
& Cycle-GAN + MT			&~ &$\surd$	            &14.4  &	17.1	 & 11.5	 &11.9   &11.1    &12.8	\\	
& Mocycle-GAN                    &$\surd$ &$\surd$                                  &\textbf{15.2}  &	\textbf{18.1}	 & \textbf{11.9}	 &\textbf{12.3}   &\textbf{11.6}    &\textbf{13.2}	\\ \Xhline{2\arrayrulewidth}
\end{tabular}
\vspace{-0.24in}
\end{table}

Figure \ref{fig:viper}(a) showcases five examples of video-to-labels results with different methods under various ambient conditions. As illustrated in the figure, our Mocycle-GAN obtains much more promising video-to-labels results. For instance, the majority categories, e.g., road (first row), cannot be well translated for baselines. Instead, even the minority classes such as car (third row) and building (fourth row) are translated nicely using our Mocycle-GAN.

\textbf{Evaluation on Labels-to-Video.} In this scenario, given an input sequence of segmentation label maps, the video translator outputs a video that resembles a real game scene video. Table \ref{table:seg2vid viper} shows the results on labels-to-video translation task on Viper. Our Mocycle-GAN performs consistently better than other methods over three metrics. Similar to the observations on the video-to-labels translation task, Recycle-GAN exhibits better performance than Cycle-GAN, by synthesising future frames via temporal predictor to explore cycle consistency across both domains and time. The further performance improvement is attained when combining Cycle-GAN and Recycle-GAN. In addition, Cycle-GAN$_{SF}$ explores motion across domains to directly constrain the temporal dynamics between synthetic frames with source motion and achieves better performances than Recycle-GAN$_{cmb}$. Furthermore, by steering unpaired video translation with the guidance from motion cycle consistency and motion translation across domains, our Mocycle-GAN boosts up the performances over all the three metrics.

Figure \ref{fig:viper}(b) shows five examples of labels-to-video results under variant ambient conditions. Clearly, our Mocycle-GAN generates more natural and vivid frames compared with the results of baselines. Concretely, our results contain more realistic objects (e.g., road, tree, and car) with plenty of details, while the other methods always generate repeated patterns and fail to capture the details.

\begin{table}[]\scriptsize
\centering
\caption{\small Ablation study for each design (i.e., Motion Cycle Consistency (MC) and Motion Translation (MT)) in Mocycle-GAN for labels-to-video on Viper.}
\setlength{\tabcolsep}{3.98pt}
\def\arraystretch{1.2}
\vspace{-0.15in}
\label{table:ablation-seg2vid}
\begin{tabular}{@{}c |l |c c |c c c c c c }
\Xhline{2\arrayrulewidth}
Criterion& Approach &MC &MT  & day  &   sunset & rain &  snow & night &  all \\
\hline\hline
\multirow{3}*{\textbf{MP}} & Cycle-GAN + MC		&$\surd$ &~     &40.3  &	58.6	 & 29.5	 &43.8   &27.9    &44.7	\\
& Cycle-GAN + MT   &~	&$\surd$            &39.0  &	57.7	 & 33.3	 &46.3   &27.7    &47.0	\\
& Mocycle-GAN                      &$\surd$ &$\surd$     &\textbf{42.1}  &	\textbf{61.2}	 & \textbf{34.6}	 &\textbf{48.1}   &\textbf{30.5}    &\textbf{47.6}	\\ \hline

\multirow{3}*{\textbf{AC}}  & Cycle-GAN + MC		 &$\surd$ &~        &14.5  &	16.3	 & 11.0	 &13.2   &14.7    &13.6	\\
& Cycle-GAN + MT		&~	&$\surd$	            &14.6  &	16.1	 & 11.3	 &13.9   &14.5    &14.5	\\	
& Mocycle-GAN                       &$\surd$ &$\surd$     &\textbf{15.4}  &	\textbf{17.6}	 & \textbf{12.6}	 &\textbf{14.9}   &\textbf{16.5}    &\textbf{16.0}	\\ \hline

\multirow{3}*{\textbf{IoU}} & Cycle-GAN + MC		 &$\surd$     &~        &9.4  &	11.0	 & 6.2	 &7.3   &7.0    &7.6	\\
& Cycle-GAN + MT		&~  &$\surd$	            &9.2  &	11.0	 & 6.5	 &8.2   &6.7    &8.6	\\ 	
& Mocycle-GAN      &$\surd$ &$\surd$       &\textbf{9.7}  &\textbf{11.9}	&\textbf{7.5}	 & \textbf{8.8}  &\textbf{7.7}    &\textbf{10.1}	\\ \Xhline{2\arrayrulewidth}

\end{tabular}
\vspace{-0.2in}
\end{table}

\textbf{Ablation Study.} In this section, we further study how each design in our Mocycle-GAN affects the overall performance. Motion Cycle consistency (\textbf{MC}) exploits the cycle consistency on motion to enforce the reconstruction of motion through translation cycle. Motion Translation (\textbf{MT}) transfers the optical flow across domains and further strengthens the temporal continuity in target domain by steering video translation with the transferred optical flow. Table \ref{table:ablation-vid2seg} and Table \ref{table:ablation-seg2vid} details the performance improvements by considering different designs for video-to-labels and labels-to-video on Viper, respectively. In particular, by further integrating motion cycle consistency and motion translation constraint into Cycle-GAN, Cycle-GAN + MC and Cycle-GAN + MT exhibit better performance than Cycle-GAN. Combining the two motion-guided temporal constraints, our Mocycle-GAN obtains the best performances on both video-to-labels and labels-to-video translations.

\begin{figure}[!tb]
    \centering {\includegraphics[width=0.43\textwidth]{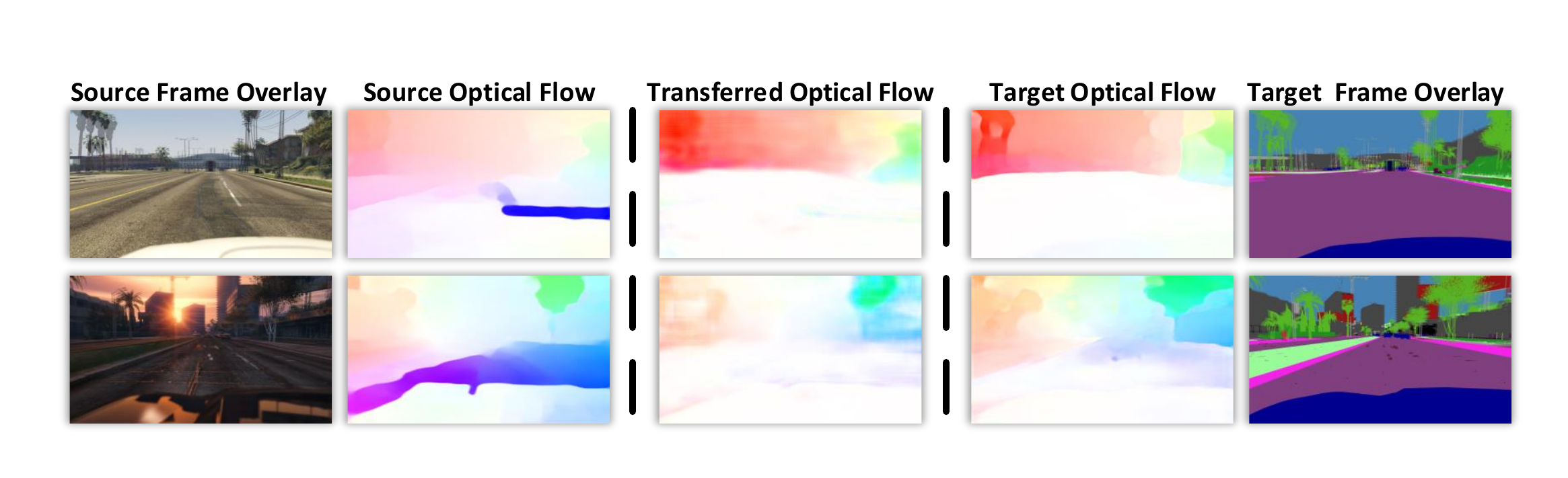}}
    \vspace{-0.253in}
    \caption{\small Examples of motion translation results in video-to-labels. From left to right: Source frame overlay, optical flow in source, transferred optical flow via motion translator, ground truth optical flow in target, and ground truth target frame overlay.}
    \label{fig:MT}
    \vspace{-0.25in}
\end{figure}

Moreover, to fully verify the effectiveness of the devised motion translation constraint, here we compare Cycle-GAN + MT against the best competitor Cycle-GAN$_{SF}$ which also exploits the motion information across domains. Unlike Cycle-GAN$_{SF}$ enforces the temporal coherence among synthetic frames in a brute-force manner, Cycle-GAN + MT elegantly transfers optical flow across domains to model the temporal coherence in target domain and thus achieves better performances. Figure \ref{fig:MT} further showcases two examples of motion translation in video-to-labels. As illustrated in the figure, the optical flows in source and target domains are substantially different, and the transferred optical flow obtained by our motion translator ends up matching closely to the ground truth optical flow in target. The results again confirms the importance of transferring motion across domains for video translation.

\begin{figure}[!tb]
\vspace{-0.05in}
    \centering {\includegraphics[width=0.44\textwidth]{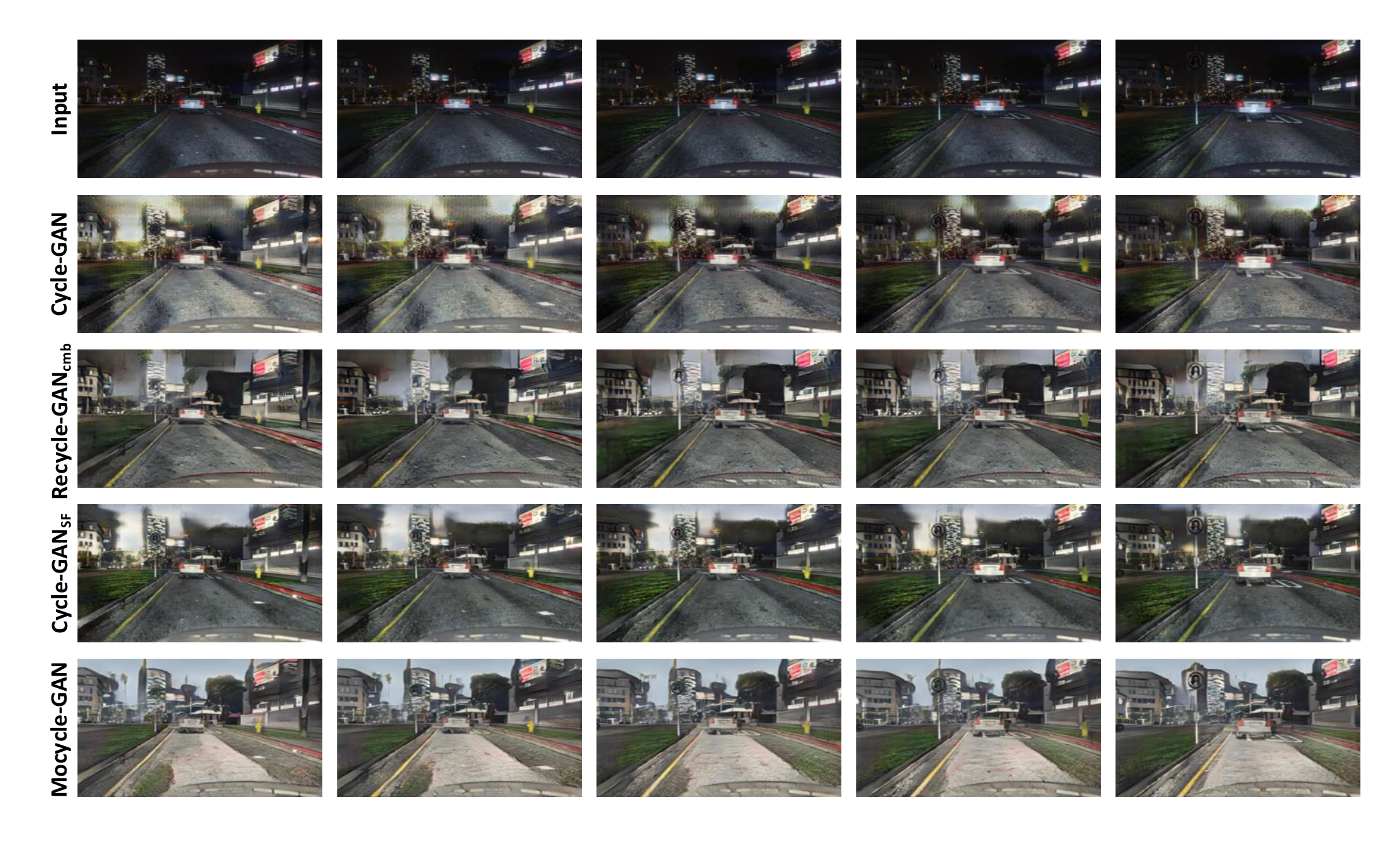}}
    \vspace{-0.28in}
    \caption{\small Examples of night-to-day results in Viper dataset. The original inputs and the output results by different models are given. Each row denotes one sequence of frames.}
    \label{fig:environment}
    \vspace{-0.3in}
\end{figure}

\subsection{Other Video Translations}

\textbf{Ambient Condition transfer.} As an universal unpaired video translator, we test our Mocycle-GAN on ambient condition transfers which explore the translation between different ambient conditions. Figure \ref{fig:environment} shows the translated videos by our Mocycle-GAN and other baselines on night-to-day task. As depicted in the figure, the baselines all generate frames whose overall color is somewhat bleak. In contrast, the color of our results gets much brighter, which better matches the style of day-time videos. Besides, our Mocycle-GAN takes the advantages of exploring both motion cycle consistency and motion translation, and thus achieves more realistic and temporal consistent videos than other methods.

\begin{figure}[!tb]
\vspace{-0.05in}
    \centering {\includegraphics[width=0.475\textwidth]{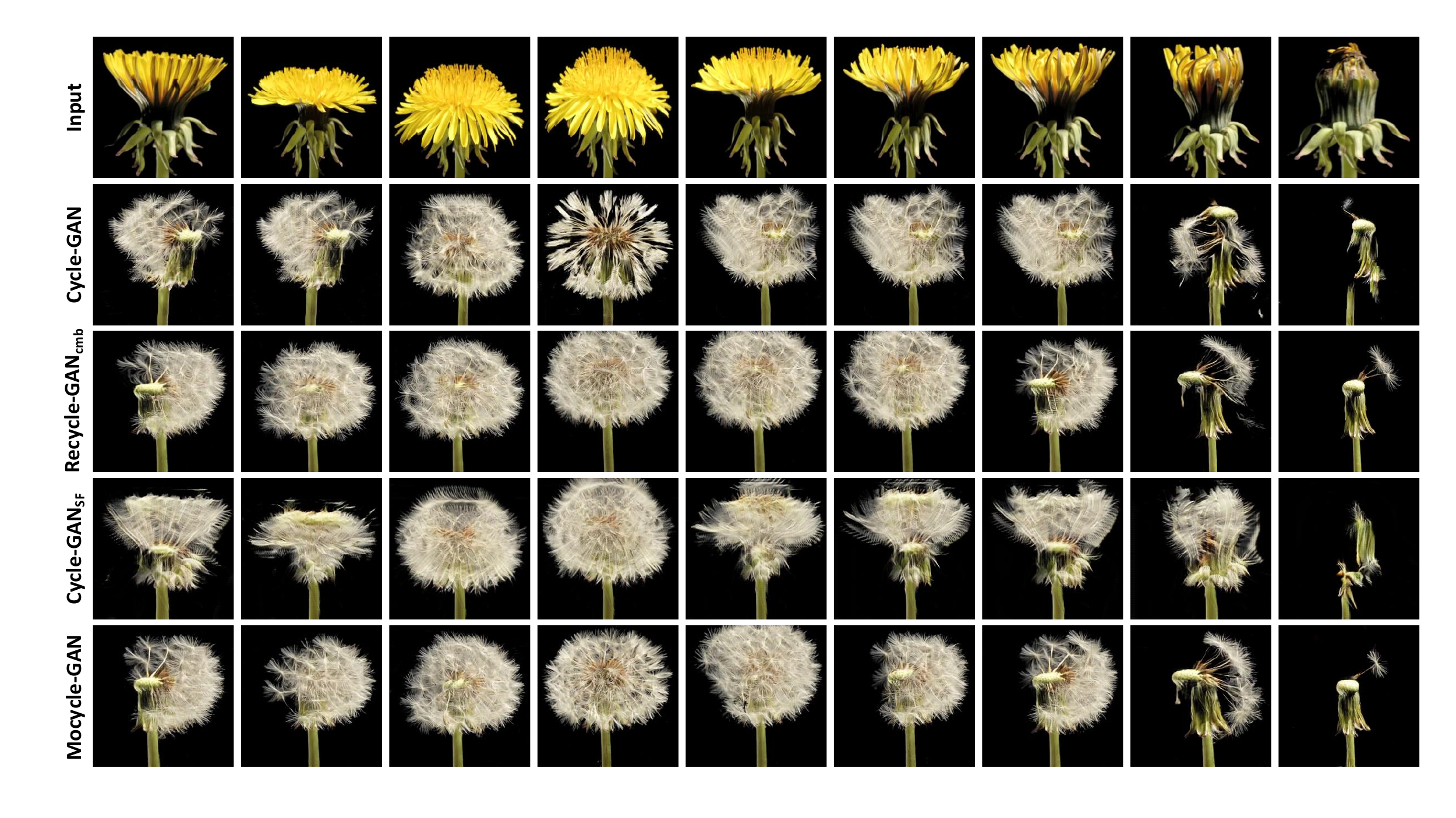}}
    \vspace{-0.435in}
    \caption{\small Examples of flower-to-flower results. The original inputs and the output results by different models are given. Each row denotes one sequence of frames.}
    \vspace{-0.18in}
    \label{fig:flowrer}
\end{figure}

\textbf{Flower-to-Flower.} We further evaluate our Mocycle-GAN on flower-to-flower that considers the translation between different flowers. The examples of translated videos by different methods are shown in Figure \ref{fig:flowrer}. Similar to the observations for ambient condition transfer, our Mocycle-GAN generates the most realistic and temporal continuous frames, where the target flower blooms and fades in synch with the source flower. This again validates the effectiveness of guiding video translation with motion information.

\textbf{Human Evaluation.}
We additionally conducted a human study to quantitatively evaluate Mocycle-GAN against three baselines, i.e., Cycle-GAN, Recycle-GAN$_{cmb}$, and Cycle-GAN$_{SF}$ on ambient condition transfer and flower-to-flower tasks. For each task, we invite 10 labelers and randomly select 80 videos clips from testing set for human evaluation. We show each input video clip with two translated results (generated by our Mocycle-GAN and one baseline) at a time and ask the labelers: which one looks more realistic and natural? According to all labelers' feedback, we measure the human preference score of one method as the percentage of its translation results that are preferred. Table \ref{table:human study} shows the results of human study. Clearly, our Mocycle-GAN is the winner on both translation tasks.

\begin{table}[]\scriptsize
\centering
\caption{\small Human preference score ($\%$) on translation quality for ambient condition transfer and flower-to-flower.}
\setlength{\tabcolsep}{5pt}
\def\arraystretch{1.2}
\vspace{-0.15in}
\label{table:human study}
\begin{tabular}{@{}l| c c  }
\Xhline{2\arrayrulewidth}
Human preference score  & Ambient Condition Transfer  &   Flower-to-Flower \\ \hline\hline
Mocycle-GAN / Cycle-GAN 	 	&\textbf{82.5} / 17.5  &	\textbf{77.5} / 22.5		\\  	
Mocycle-GAN / Recycle-GAN$_{cmb}$   &\textbf{73.8} / 26.2  &	\textbf{72.5} / 27.5	\\
Mocycle-GAN / Cycle-GAN$_{SF}$  &\textbf{66.3} / 33.7  &\textbf{88.8} / 11.2   \\
\Xhline{2\arrayrulewidth}
\end{tabular}
\vspace{-0.23in}
\end{table}

\section{Conclusions}
We have presented Motion-guided Cycle GAN (Mocycle-GAN) architecture, which explores both appearance structure and temporal continuity for video-to-video translation in an unsupervised manner. In particular, we study the problem from the viewpoint of integrating motion estimation into unpaired video translator. To verify our claim, we devise three types of spatial/temporal constrains: adversarial constraint is to discriminate between synthetic and real frames in an adversarial manner and thus enforce each synthetic frame realistic at appearance; frame and motion cycle consistency constraints encourage the reconstruction of both appearance structure in frames and temporal continuity in motion; motion translation constraint validates the transfer of motion across domains which further strengthens the temporal continuity. Extensive experiments conducted on video-to-labels and labels-to-video translation validate our proposal and analysis. More remarkably, the qualitative results and human study on more translations, e.g., flower-to-flower and ambient condition transfer, demonstrate the efficacy of Mocycle-GAN.

\textbf{Acknowledgments.}
This work was supported in part by NSFC projects 61872329 and 61572451.
\bibliographystyle{ACM-Reference-Format}
\bibliography{sample-sigconf}

\end{document}